\documentclass[sigconf]{acmart}
\copyrightyear{2023}
\acmYear{2023}
\setcopyright{acmlicensed}\acmConference[CIKM '23]{Proceedings of the 32nd ACM International Conference on Information and Knowledge Management}{October 21--25, 2023}{Birmingham, United Kingdom}
\acmBooktitle{Proceedings of the 32nd ACM International Conference on Information and Knowledge Management (CIKM '23), October 21--25, 2023, Birmingham, United Kingdom}
\acmPrice{15.00}
\acmDOI{XXXXXXX.XXXXXXX}
\acmISBN{979-8-4007-XXXX-X/23/10}

\usepackage{graphicx}
\usepackage{amsfonts} 
\usepackage{amsmath}
\usepackage{multirow} 
\usepackage{makecell}
\usepackage{subfigure} 
\usepackage{booktabs} 
\usepackage{makecell}
\usepackage{pifont}

\begin{document}


\title[\textsc{GripRank:} Bridging the Gap between Retrieval and Generation via the \\  Generative Knowledge Improved Passage Ranking]{\textsc{GripRank:} Bridging the Gap between Retrieval and Generation via the Generative Knowledge Improved \\ Passage Ranking}

\author{Jiaqi Bai}
\affiliation{%
  \institution{School of Cyber Science and Technology, Beihang University}
  \city{Beijing}
  \country{China}
}
\email{bjq@buaa.edu.cn}

\author{Hongcheng Guo}
\affiliation{
  \institution{State Key Lab of Software Development Environment, \\ Beihang University}
  \city{Beijing}
  \country{China}
  }
\email{hongchengguo@buaa.edu.cn}

\author{Jiaheng Liu}
\affiliation{
  \institution{State Key Lab of Software Development Environment, \\ Beihang University}
  \city{Beijing}
  \country{China}
  }
\email{liujiaheng@buaa.edu.cn}

\author{Jian Yang}
\affiliation{
  \institution{DAMO Academy, \\ Alibaba Group}
  \city{Beijing}
  \country{China}
  }
\email{jiaya@buaa.edu.cn}

\author{Xinnian Liang}
\affiliation{
  \institution{State Key Lab of Software Development Environment,\\ Beihang University}
  \city{Beijing}
  \country{China}
  }
\email{xnliang@buaa.edu.cn}

\author{Zhao Yan}
\affiliation{
  \institution{Tencent Cloud AI}
  \city{Beijing}
  \country{China}
  }
\email{zhaoyan@tencent.com}

\author{Zhoujun Li}
\authornote{Corresponding Author}
\affiliation{
  \institution{State Key Lab of Software Development Environment, \\ Beihang University}
  \city{Beijing}
  \country{China}
  }
\email{lizj@buaa.edu.cn}

\renewcommand{\shortauthors}{Jiaqi Bai et al.}

\begin{abstract}

Retrieval-enhanced text generation has shown remarkable progress on knowledge-intensive language tasks, such as open-domain question answering and knowledge-enhanced dialogue generation, by leveraging passages retrieved from a large passage corpus for delivering a proper answer given the input query.
However, the retrieved passages are not ideal for guiding answer generation because of the discrepancy between retrieval and generation, 
i.e., the candidate passages are all treated equally during the retrieval procedure without considering their potential to generate a proper answer.
This discrepancy makes a passage retriever deliver a sub-optimal collection of candidate passages to generate the answer.
In this paper, we propose the \textbf{G}ene\textbf{R}ative Knowledge \textbf{I}mproved \textbf{P}assage \textbf{Rank}ing (\textbf{\textsc{GripRank}}) approach, addressing the above challenge by distilling knowledge from a generative passage estimator (GPE) to a passage ranker, where the GPE is a generative language model used to measure how likely the candidate passages can generate the proper answer.
We realize the distillation procedure by teaching the passage ranker learning to rank the passages ordered by the GPE.
Furthermore, we improve the distillation quality by devising a curriculum knowledge distillation mechanism, which allows the knowledge provided by the GPE can be progressively distilled to the ranker through an easy-to-hard curriculum, enabling the passage ranker to correctly recognize the provenance of the answer from many plausible candidates. 
We conduct extensive experiments on four datasets across three knowledge-intensive language tasks.
Experimental results show advantages over the state-of-the-art methods for both passage ranking and answer generation on the KILT benchmark.

\end{abstract}



\maketitle

\section{Introduction}

Knowledge-intensive language tasks, including open-domain question answering, knowledge-grounded conversation, and fact verification, pose a challenge for retrieving passages most likely to be the provenance of the target answer from a large passage corpus (e.g., Wikipedia). 
One of the most successful paradigms to deal with these tasks is retrieval-enhanced text generation \cite{DBLP:conf/nips/LewisPPPKGKLYR020,DBLP:journals/corr/abs-2002-08909,DBLP:conf/eacl/IzacardG21,DBLP:journals/corr/abs-2211-12561}.
It first trains a passage retriever (e.g., DPR \cite{DBLP:conf/emnlp/KarpukhinOMLWEC20} and GTR \cite{DBLP:conf/emnlp/Ni0LDAMZLHCY22}) to collect passages relevant to the input query.
Then, it uses a generative language model (e.g., BART \cite{DBLP:conf/acl/LewisLGGMLSZ20} and T5 \cite{DBLP:journals/jmlr/RaffelSRLNMZLL20}) to generate the answer grounding on the retrieved passages and the input query.

\begin{figure}[t]
\centering
\includegraphics[width=8.0cm,height=5.8cm]{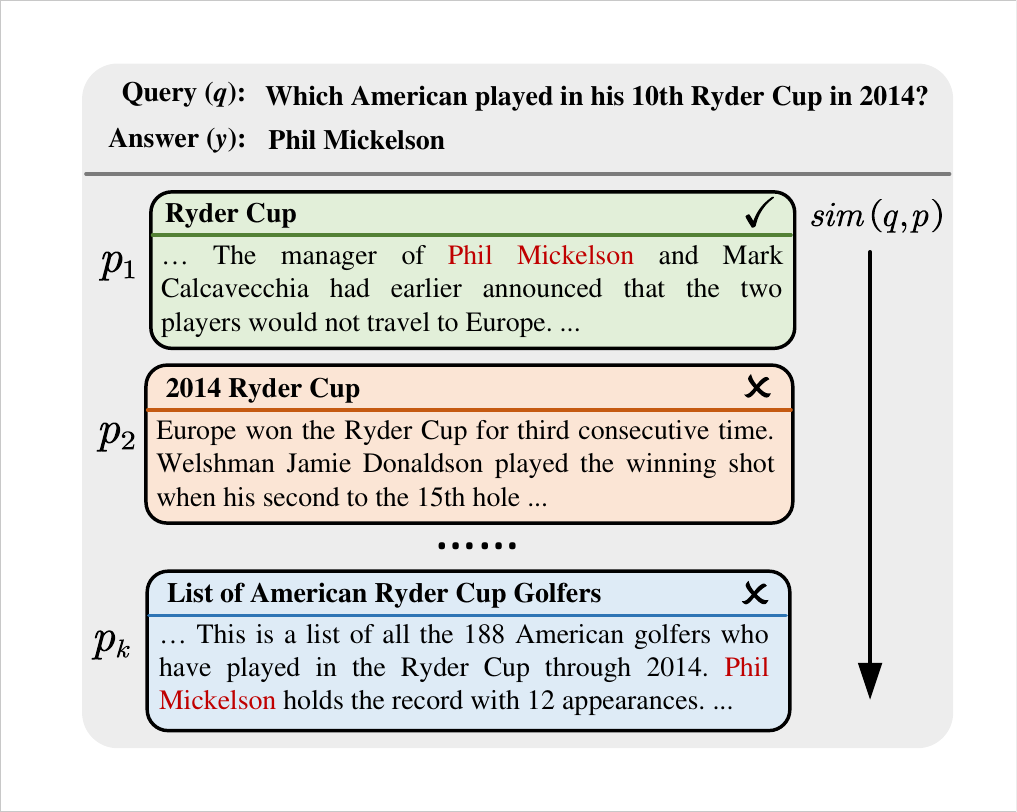}
\caption{An example taken from the TriviaQA dataset \cite{DBLP:conf/acl/JoshiCWZ17} standardized by the KILT benchmark \cite{DBLP:conf/naacl/PetroniPFLYCTJK21}. We show that the golden answer hides in the passage which is not highly relevant to the input query. The $sim( q,p )$ with the down arrow denotes the passages are ranked in descending order according to the similarity with the input query $q$. The ``\ding{52}'' and ``\ding{56}'' denote the passage is annotated as golden and non-golden for answer generation, respectively.}
\label{fig_intro}
\vspace{-3mm}
\end{figure}

Ideally, a passage retriever with good retrieval capability can provide passages more likely to be the provenance of the golden answer.
Some efforts have been made toward this goal by improving the relevance of the retrieved passages to the input query. 
For example, mining the candidate passages to serve as hard negatives \cite{DBLP:conf/emnlp/KarpukhinOMLWEC20,DBLP:journals/corr/abs-2102-02096, DBLP:conf/emnlp/GlassRCG21,DBLP:conf/sigir/ZhanM0G0M21} or leveraging a passage ranker to improve the ranking quality of the candidates selected by the passage retriever
\cite{DBLP:conf/naacl/GlassRCNCG22,DBLP:conf/emnlp/0003GJQ00ZCYD22,DBLP:journals/corr/abs-2209-14290,DBLP:conf/sigir/FanLZLLX22}.
These methods demonstrate the importance of enhancing the relevancy between the input query and the selected candidates from different perspectives.
However, they ignore that the golden answer may hide in the passages which are not highly relevant to the input query.
As shown in Figure \ref{fig_intro}, we present the passages retrieved by DPR according to the input query, sorting in descending order regarding the predicted similarity score.
In this case, the golden passage $p_1$ is correctly retrieved at the top from the passage corpus, while another reasonable passage $p_k$ that contains the golden answer is ranked at the tail of the candidate passage pool, which will be ignored during the answer generation.
One of the main reasons for the above challenge is that the candidate passages are all treated equally during the retrieval procedure without considering their potential to generate the golden answer. 
Moreover, the deficiency of such critical knowledge will result in the retriever being hard to correctly recognize the provenance of the golden answer from many plausible candidate passages.

Some pioneering work tried to handle the above challenges. 
A solution that would at first glance seem obvious is end-to-end training of the passage retriever and the answer generator \cite{DBLP:conf/nips/LewisPPPKGKLYR020,DBLP:journals/corr/abs-2002-08909}, in which both components are optimized to generate the correct answer regardless of the golden passage label.
However, this resolution lacks a mechanism to explicitly measure how likely the selected passages can be used to generate the correct answer, which may damage the performance of the passage retriever when there are too many plausible candidates to distinguish provenance from them correctly \cite{DBLP:conf/acl/0001G20}, resulting in the golden passage being ignored during the answer generation.
Another line of work approaches the posterior information during the passage retrieval procedure\cite{DBLP:conf/emnlp/RenQLZSWWW21,DBLP:conf/iclr/ParanjapeKPZM22,DBLP:conf/www/LinGLZLD0LJMD23}, where the posterior information can be incorporated into a teacher model like a pre-trained passage retriever or a passage ranker.
However, this line of work only considers the posterior information from the encoder part, which lacks a mechanism to explicitly measure how likely the retrieved passages can be used to generate the golden answer autoregressively.

In this paper, we propose the \textbf{G}ene\textbf{R}ative Knowledge \textbf{I}mproved \textbf{P}assage \textbf{Rank}ing (\textbf{\textsc{GripRank}}) approach by distilling knowledge from a generative passage estimator (GPE) to a passage ranker. 
The GPE is a generative language model \cite{DBLP:conf/acl/LewisLGGMLSZ20,DBLP:conf/acl/YangM0HHYZYWL23}, 
which explicitly measures how likely a candidate passage can be used to guide the generation of the golden answer.
We train a GPE by feeding the concatenation of the query and the golden passage, generating the golden answer autoregressively.
Once the training is finished, we freeze the entire GPE, taking the concatenation of the query and the candidate passage as input, leveraging the cross-entropy loss between the predicted and the golden answer to measure how likely a candidate passage can be the provenance.
The distillation procedure can be realized by teaching a student learning to rank the passages ordered by the GPE. 
Compared to the work that distills knowledge from the encoder-only architecture \cite{DBLP:conf/emnlp/RenQLZSWWW21,DBLP:conf/iclr/ParanjapeKPZM22}, the generative framework provides fine-grained knowledge to measure the relevance between the candidate passages and the golden answer in an autoregressive manner.
However, there can be one-to-many relations between the golden answer and the selected passages in the real-world scenario \cite{DBLP:conf/acl/JoshiCWZ17,dinan2018wizard,kwiatkowski-etal-2019-natural}, especially when the passage corpus is large (e.g., Wikipedia), which may result in the golden passage out-of-recall in the top-ranked candidates.
To enforce that the golden passage always ranks at the top of the output distribution estimated by the GPE, we devise a label rectification method by introducing a balanced term between the golden passage label and the output distribution.
Furthermore, to make the ranker better distinguish the provenances of the golden answer from many plausible candidates, 
we devise a novel curriculum knowledge distillation mechanism to progressively distill knowledge from the GPE to the ranker, which is realized by dynamically controlling the difficulty of the sampled passages from the candidate passage pool through an easy-to-hard curriculum \cite{DBLP:conf/icml/BengioLCW09}.
We conduct experiments on four datasets across three tasks standardized by the KILT benchmark, including ZSRE \cite{DBLP:conf/conll/LevySCZ17} (zero-shot slot filling task), TriviaQA \cite{DBLP:conf/acl/JoshiCWZ17} and Natural Questions \cite{kwiatkowski-etal-2019-natural} (open-domain question answering task), and Wizard of Wikipedia \cite{dinan2018wizard} (Knowledge-enhanced dialogue generation task).
Experimental results show the superior performance of the proposed \textsc{GripRank} over the previous state-of-the-art for both passage ranking and answer generation.
Further analysis demonstrates the effectiveness of the proposed approach.

Our contributions are three-fold:
\textbf{i)}: We introduce \textsc{GripRank}, the first generative knowledge improved passage ranking approach for improving the answer generation in knowledge-intensive language tasks.
\textbf{ii)}: We propose a novel curriculum knowledge distillation mechanism for better distilling the generative knowledge to a passage ranker.
\textbf{iii)}: Extensive experiments and thorough analysis demonstrate the effectiveness of our proposed approach. We present that distilling knowledge from a generative passage estimator to a passage ranker effectively narrows the gap between passage retrieval and answer generation.
\section{Related Work}

\subsection{Knowledge-Intensive Language Tasks}

Knowledge-Intensive Language Tasks (KILT) are a collection of tasks that require a system to produce a proper answer according to the input query by accessing external knowledge sources (e.g., Wikipedia).
For example, the zero-shot slot-filling task \cite{DBLP:conf/conll/LevySCZ17} aims to recognize a set of relations for a given entity and use them to populate structures \cite{DBLP:conf/emnlp/GlassRCG21}.
Open-domain question answering \cite{DBLP:conf/acl/JoshiCWZ17,yang-etal-2018-hotpotqa,kwiatkowski-etal-2019-natural,fan-etal-2019-eli5} aims to produce the correct answer for a given question by accurately locating the provenance from knowledge sources.
Knowledge-enhanced dialogue generation \cite{dinan2018wizard,DBLP:journals/corr/abs-2306-15430} aims to generate the proper response for the given dialogue history by considering knowledge relevant to the dialogue context.
 
There is plenty of datasets have been proposed for KILT tasks \cite{DBLP:conf/conll/LevySCZ17,elsahar:hal-01906329,10.3233/SW-170273}.
While these datasets always have different input formats and evaluation schemes.
Moreover, they always access knowledge from different knowledge sources.
To facilitate the comparison among these tasks, a benchmark named KILT \cite{DBLP:conf/naacl/PetroniPFLYCTJK21} was introduced to unify these tasks.
All tasks in KILT are formulated into a unified interface and grounded in the same snapshot of Wikipedia.
In this work, we focus on three tasks standardized by the KILT benchmark, including zero-shot slot-filling, open-domain question answering, and knowledge-enhanced dialogue generation.

\subsection{Retrieval-Enhanced Text Generation}
Retrieval-enhanced text generation aims to produce a proper answer for the input query by leveraging a retrieval component to collect evidence from large knowledge sources, which achieves competitive performance across many knowledge-intensive language tasks \cite{maillard-etal-2021-multi,DBLP:journals/corr/abs-2112-09924,NEURIPS2022_cd88d62a,DBLP:conf/naacl/GlassRCNCG22,DBLP:journals/corr/abs-2209-14290}.
It has been demonstrated that the passage retrieval performance significantly impacts the quality of the final generated answer.
Following this intuition, researchers attempt to incorporate the posterior information into the passage retrieval procedure \cite{DBLP:conf/iclr/ParanjapeKPZM22,DBLP:conf/www/LinGLZLD0LJMD23,DBLP:journals/corr/abs-2301-12652} or leverage a passage ranker to further boost the ranking quality of the candidates retrieved by a passage retriever \cite{fajcik-etal-2021-r2-d2,DBLP:conf/emnlp/RenQLZSWWW21,DBLP:conf/naacl/GlassRCNCG22}.
For example, \citet{DBLP:conf/emnlp/RenQLZSWWW21} introduced a joint training approach to jointly optimize the retriever and re-ranker by distilling knowledge from each other.
\citet{DBLP:conf/naacl/GlassRCNCG22} employed a passage ranker with cross-encoder architecture to improve the passage retrieval capability and achieved state-of-the-art results on the KILT benchmark.
\citet{DBLP:conf/iclr/ParanjapeKPZM22} incorporated the posterior information into the passage retrieval procedure. 
They leverage an additional passage retriever as a teacher, which takes the golden answer as input and uses the output distribution of the teacher to supervise the learning of the student.

We follow the line of research that uses the passage ranker to enhance the ranking quality of the collected candidate passages by passage retriever.
However, we focus on leveraging a passage ranker to mitigate the discrepancy between retrieval and generation by distilling the knowledge from a generative language model.

\subsection{Knowledge Distillation}
Knowledge distillation \cite{DBLP:journals/corr/HintonVD15} aims to transfer knowledge from a stronger pre-trained teacher model to a compact student model. 
One of the key ideas of knowledge distillation is to inject the posterior knowledge (e.g., the provenance of the golden answer) into the teacher, and the student is supervised by matching the output distribution of the teacher during the training procedure. 

Conventional methods conduct knowledge distillation by minimizing the Kullback-Leibler divergence loss (KLD loss) between the output distribution of the teacher model and the student model \cite{DBLP:journals/corr/HintonVD15,DBLP:journals/corr/abs-1909-10351,DBLP:conf/acl/ChenGCLL20,um4}. 
Recent studies treat knowledge distillation as a learning-to-rank problem \cite{DBLP:conf/icml/CaoQLTL07,DBLP:conf/aistats/ReddiPMRYKVK21,NEURIPS2022_aa31dc84}.
Thus, the distillation process can be regarded as a teacher teaching a student how to order the top-ranked candidates correctly.
Another line of work improves the distillation performance by progressively distilling knowledge from a teacher model to a student model.
For example, \citet{DBLP:conf/www/LinGLZLD0LJMD23} proposed a progressive distillation approach for dense passage retrieval, which gradually improves the capability of teachers and students through both model and data perspectives.
\citet{DBLP:journals/corr/abs-2211-16231} enhanced the distillation performance by progressively increasing the learning difficulty of the student.
They realized the progressive distillation by controlling the distillation temperature \cite{DBLP:journals/corr/HintonVD15} in an easy-to-hard curriculum \cite{DBLP:conf/icml/BengioLCW09,DBLP:journals/taslp/BaiYYGL23}, where the distillation temperature is a hyper-parameter that controls the smoothness of probability distributions and can faithfully determine the difficulty level of the distillation process \cite{DBLP:conf/icml/ChandrasegaranT22,DBLP:journals/corr/abs-2202-07940}.

In contrast to the above work, we train a generative passage estimator as a teacher and distill the ranking order estimated from it to the student, which allows the student to learn how to order the top-ranked passages.
Moreover, we realize the knowledge distillation in an easy-to-hard manner by progressively increasing the difficulty of the sampled candidate passages for training the student, which enables the student progressively learn to distinguish the provenance of answers from many plausible candidates.

\section{Methodology}

\begin{figure*}[t]
\centering
\includegraphics[width=18.cm]{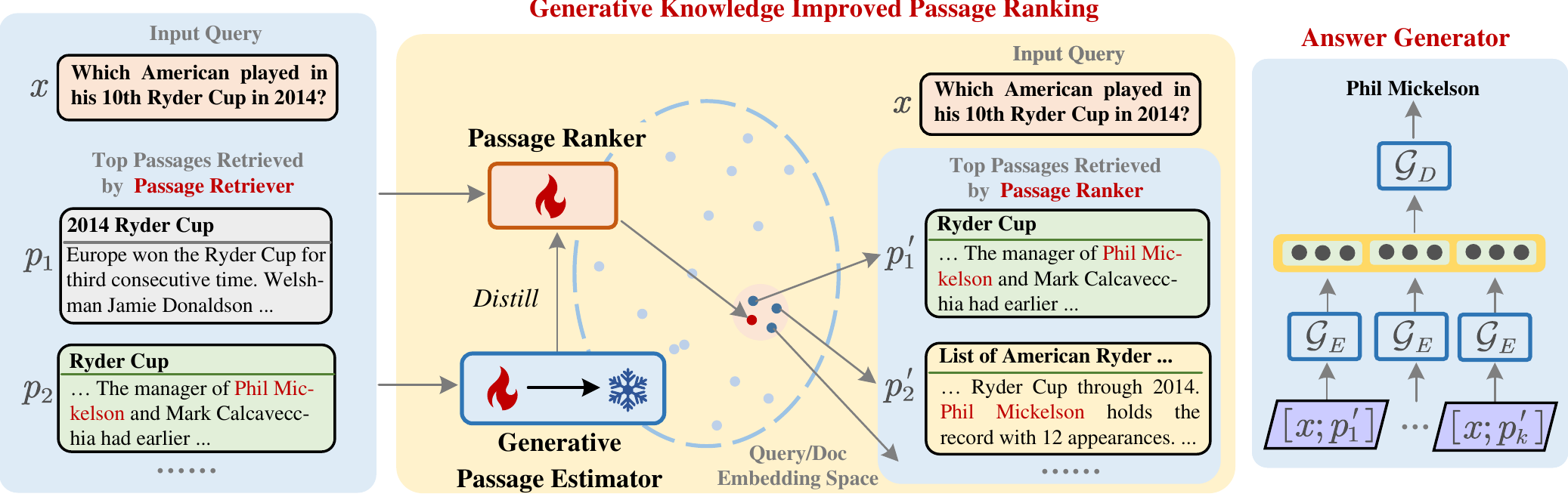}
\caption{The architecture of our proposed \textsc{GripRank}. We train a generative passage estimator (GPE) under the supervision of the golden answer, taking the concatenation of the query and golden passage as input. Once the GPE finishes training, we freeze the entire GPE and the passage ranker learns to rank the candidate passages ordered by the GPE. Each top-ranked passage is then concatenated with the input query for answer generation.  }
\label{fig_main}
\end{figure*}

\subsection{Problem Formalization}
Knowledge-intensive NLP tasks aim to generate answer $y$ given input query $x$ with the usage of a list of retrieved passages $\mathbf{P}$, where $\mathbf{P}=\{p_1,...,p_{\left| \mathbf{P} \right|}\}$ is related to the input query $x$ and retrieved from a large passage corpus such as Wikipedia.

A popular solution to deal with these tasks employs a retrieval-enhanced text generation framework (e.g., FiD \cite{DBLP:conf/eacl/IzacardG21}, RAG \cite{DBLP:conf/nips/LewisPPPKGKLYR020} and REALM \cite{DBLP:journals/corr/abs-2002-08909}). 
These architectures typically consist of two components: a passage retriever $\mathcal{Q}$ and an answer generator $\mathcal{G}$.
The passage retriever $\mathcal{Q}(\mathbf{P}|x;\theta _{\mathcal{Q}})$ with parameters $\theta _{\mathcal{Q}}$ is trained to retrieve a list of passages $\mathbf{P}=\{p_1,...,p_{| \mathbf{P} |}\}$ with the $|\mathbf{P}|$ most relevant scores for input query $x$.
The answer generator $\mathcal{G}(y_i|x,\mathbf{P},y_{1:i-1};\theta _{\mathcal{G}})$ parameterized by $\theta _{\mathcal{G}}$ is then trained to generate answer $y$ token by token, given the input query $x$ and retrieved passages $\mathbf{P}$.

Usually, the relevance between the retrieved passages and the ground-truth answer significantly impacts the correctness of the generated results.
Because the more passage relevant to the input query, the more likely it contains the provenance of the answer.
With this intuition, we aim to leverage a passage ranker $\mathcal{R}(\mathbf{P}^{\prime}|x,\mathbf{P};\theta _{\mathcal{R}})$
parameterized by $\theta _{\mathcal{R}}$, to improve the ranking order of passages in $\mathbf{P}$ given input query $x$, gathering a list of re-ordered passages $\mathbf{P}^{\prime}=\{p_{1}^{\prime},...,p_{| \mathbf{P} |}^{\prime}\}$ for answer generation.

\subsection{Method Overview}
We improve the conventional retrieval augmented generation framework by introducing a ranker to enhance the ranking order of passages collected from the passage retriever. Thus, a passage retriever is trained to first retrieve candidate passages according to the given input query. Then, the ranker further ranks retrieved candidates, exploring the most relevant passages to the golden answer.

The architecture of our proposed method is shown in Figure \ref{fig_main}.
Different from previous work \cite{DBLP:conf/emnlp/RenQLZSWWW21,DBLP:conf/naacl/GlassRCNCG22}, we improve the passage ranker by distilling knowledge from a generative passage estimator (GPE), which is trained to estimate how likely the provided candidate passages can generate the ground-truth answer by combining the input query.
Once the distillation process is finished, the ranker re-ranks the candidate passages retrieved from the passage retriever, aiming to collect the most relevant passages for generating the golden answer.

\subsection{Passage Ranker}
Given an input query $x$, and a list of candidate passages $\mathbf{P}$ retrieved by the passage retriever $\mathcal{Q}$, a vanilla passage ranker aims to enhance the relevance level of a candidate passage $p_k$ ($p_k\in \mathbf{P}$) to the input query $x$.
We implement the passage ranker with cross-encoder architecture \cite{DBLP:journals/corr/abs-1904-07531,DBLP:journals/corr/abs-1901-04085}, as it can better capture the semantic interactions between the passage and the query.
To compute the relevance score $z_k \propto sim( x,p_k )$, we concatenate $x$ and $p_k$ separated by a special token \texttt{[SEP]}, prepending a \texttt{[CLS]} token to the beginning of the sequence.
The $z_k$ can be collected by the output representation at the \texttt{[CLS]} token encoded by $\mathcal{R}$.
Generally, we use the negative of the summed log-likelihood for the relevance score 
of the golden passage $z_g$ as the loss function:
\begin{equation}
\label{nll_loss}
\mathcal{L}_{NLL}=-\log \frac{e^{\left( z_g \right)}}{\sum\nolimits_j^{\left| \mathbf{P} \right|}{e^{\left( z_j \right)}}}
\end{equation}

\subsection{Improving Passage Ranker with Generative Passage Estimator}

Different from the previous work \cite{DBLP:conf/emnlp/RenQLZSWWW21,DBLP:conf/naacl/GlassRCNCG22} only employs the ranker to enhance the relevance between the retrieved passages and the source input. 
We improve the passage ranker by distilling the knowledge from a generative passage estimator (GPE), where the GPE is trained to measure how likely the retrieved passages can be used to generate a correct answer autoregressively.

\subsubsection{Generative Passage Estimator}

The GPE can be arbitrary generative language models, such as BART \cite{DBLP:conf/acl/LewisLGGMLSZ20} and T5 
\cite{DBLP:journals/jmlr/RaffelSRLNMZLL20}.
We train a GPE by taking the concatenation of input query $x$ and the golden passage $p_g$ as input, generating the golden answer $y$ token-by-token.
Once the training is finished, we freeze the parameters of the GPE and ask it to generate golden answer $y$ given the input query $x$ concatenated with a candidate passage $p_k$ ($p_k\in \mathbf{P}$).
We employ the normalized sentence-level cross-entropy loss of $y$ to measure how likely the candidate passage $p_k$ can be used to generate the golden answer $y$, which is formalized as follows:
\begin{gather}
z_k=\frac{\sum_j^{\left| \mathbf{P} \right|}{\sum_t^{\left| y \right|}{\log p\left( \left. y_t \right|y_{<t},x,\hat{p}_j \right)}}}{\sum_t^{\left| y \right|}{\log p\left( \left. y_t \right|y_{<t},x,\hat{p}_k \right)}} \\
r_k=\frac{e^{\left( z_k \right)}}{\sum\nolimits_j^{\left| \mathbf{P} \right|}{e^{\left( z_j \right)}}}
\end{gather}

\noindent Intuitively, a large value of $r_k$  (i.e., lower cross-entropy loss for candidate passage $\hat{p}_k$) indicates that the candidate passage $\hat{p}_k$ is more likely to be a provenance of the golden answer.

In practice, the sorting results of GPE may result in the golden passage out-of-recall in the top-ranked candidates.
Because there exist some reasonable potential resolutions which are not labeled as golden passages when the passage corpus is large.
To overcome this issue, we devise a \emph{label rectification} method to enforce the golden passage always ranked at the top of the output distribution estimated by the GPE:
\begin{equation}
r\gets \varepsilon \cdot q+\left( 1-\varepsilon \right) \cdot r
\end{equation}

\noindent where $\varepsilon$ is a rectified term to balance the probability distribution $r$ predicted by GPE and the one-hot label $q$ labeled by human annotators.
To enable the golden passage $p_g$ always receive the largest probability in $r$,  one reasonable solution is to ensure $
\varepsilon = ( 1-\varepsilon ) \max ( r_{\ne q_{p_g}} )$, where $\max ( r_{\ne q_{p_g}} )$ is the maximum probability for non-golden passages in $r$.
Thus, we have to enforce the $\varepsilon$ meets the following equation:
\begin{equation}
\varepsilon =\frac{\max \left( r_{\ne q_{p_g}} \right)}{1+\max \left( r_{\ne q_{p_g}} \right)}
\end{equation}

\subsubsection{Distilling Knowledge from the Passage Estimator to the Passage Ranker}

\noindent During training, the ranker is supervised by the estimation results of the trained GPE.
Therefore, the order of the candidate passages measured by GPE serves as a teacher and the ranker is a student. 
We employ ListMLE loss \cite{DBLP:conf/icml/XiaLWZL08} to enforce the ranker learning the ranking order from GPE: 
\begin{equation}
\label{listmle_loss}
\mathcal{L}_{ListMLE}=-\sum_{k=1}^{\left| \mathbf{P} \right|}{\log \frac{e^{\left( r_{o_k} \right)}}{\sum\nolimits_{i=k}^{\left| \mathbf{P} \right|}{e^{\left( r_{o_i} \right)}}}}
\end{equation}

\noindent where $o_i$ is the relevance order of the $i$-th candidate passage measured by GPE. 
The overall loss to optimize the ranker is the sum of the NLL loss (equation \ref{nll_loss}) and the ListMLE loss formulated as:
\begin{equation}
\mathcal{L}=\mathcal{L}_{NLL}+\mathcal{L}_{ListMLE}
\end{equation}

\noindent Once the ranker is finished training, we re-rank the candidate passages in $\mathbf{P}$ sorted by the passage retriever and output them into the re-ordered candidate sentence pool $\mathbf{P}^\prime$.

\subsubsection{Curriculum Knowledge Distillation}

The selection of candidate passages has a significant impact on the distillation quality for a passage ranker.
One of the alternative approaches is to mind candidate passages serving as hard negatives \cite{DBLP:conf/emnlp/KarpukhinOMLWEC20,DBLP:conf/emnlp/GlassRCG21,DBLP:conf/naacl/GlassRCNCG22} for improving the ranking capability.
However, in human education, teachers always train students starting with basic (easy) curricula, and then progressively exposure to more advanced (hard) curricula along with students growing up.
Students will learn much better if the curricula are organized in a reasonable order.

Inspired by curriculum learning \cite{DBLP:conf/icml/BengioLCW09}, we propose a curriculum knowledge distillation mechanism for better distilling the knowledge from GPE to ranker.
Concretely, we devise a \emph{curriculum passage sampling} strategy to realize the knowledge distillation in a curriculum manner, which is achieved by gradually increasing the difficulty of the sampled candidate passages during the distillation procedure.
Thus, it challenges the passage ranker to distinguish and order the passages in an easy-to-hard curriculum progressively.

\paragraph{Difficulty Measurement}
We define the difficulty of a candidate passage $p_j$ ($j\ne g$, $p_j\in \mathbf{P}$) as its similarity to the input query $x$, where the similarity is the normalized dot-product score predicted by the pre-trained passage retriever $\mathcal{Q}$.
We sort the passages according to the similarity in ascending order, obtaining a sorted passage pool $\mathbf{P}_{sort}=\{ \tilde{p}_1,\cdots \tilde{p}_{| \mathbf{P}_{sort} |} \}$.
Thus, a higher rank of passage means it is more similar to the input query $x$, indicating that it is more difficult for the passage ranker to distinguish it from the golden provenance $p_{g}$.

\paragraph{Curriculum Passage Sampling}

To realize curriculum knowledge distillation, we sample passages from the sorted passage pool $\mathbf{P}_{sort}$ by progressively increasing the difficulty of sampled candidates.
Concretely, we adjust the scale $\varphi$ ($\varphi \leqslant | \mathbf{P}_{sort} |$) of the sampling space $\rho _{sort}$ from which the candidate passages are sampled, where $\rho _{sort}=\{\tilde{p}_1,\cdots ,\tilde{p}_{\varphi}\}$ is a subset of $\mathbf{P}_{sort}$ (i.e., $\rho _{sort}\in \mathbf{P}_{sort}$).
The sampling scale $\varphi$ can be determined as follows:
\begin{equation}
\label{p_phi}
\varphi = 
\left\{
\begin{array}{lcr}
N_{0}
& t\le T_{0} \\
& \\
N_0+\lfloor \frac{t-T_0}{T-T_0}(\left| \mathbf{P}_{sort} \right|-N_0) \rfloor 
& T_0<t\le T \\
& \\
\left| \mathbf{P}_{sort} \right|
& t>T\\
\end{array} 
\right. 
\end{equation}

\noindent where $N_0$ ($N_0\in \mathbb{N}$, $N_0\leqslant | \mathbf{P}_{sort} |$) is the scale of warm-up passage pool.
$T_0$ and $T$ are warm-up steps and total curriculum steps, respectively. 
$\lfloor \cdot \rfloor$ denotes the fraction rounded down.

Intuitively, when the current training steps $t$ satisfy $t\le T_{0}$, the candidate passages are sampled from the set $\{\tilde{p}_1,\cdots ,\tilde{p}_{N_0}\}$.
Thus, we only use $N_0$ easiest candidates for knowledge distillation to warm up the ranker.
Once the warm-up stage is finished, we progressively expand the sampling scale $\varphi$ with the training process going on ($T_0<t\le T$).
The ranker is exposed to harder candidates, requiring a strong ranking capability for distinguishing the provenance from each plausible candidate.
When $t > T$, the ranker approaches the entire passage pool $\mathbf{P}_{sort}$ and randomly samples the candidates in $\mathbf{P}_{sort}$ to distill knowledge from GPE to ranker.

\subsection{Ranker-Augmented Answer Generator}

During answer generation, we first select the $k$ most relevant passages ranked at the top by the passage ranker. 
Then, we concatenate the input query $x$ with each selected passage $p_{m}^{\prime}$ ($m \leqslant k$) and independently feed them into the encoder of the generator ($\mathcal{G}_E$), similar to the implementation of FiD \cite{DBLP:conf/eacl/IzacardG21},  formulating as:
\begin{equation}
e_m=\mathcal{G}_E\left( \left[ \texttt{query}: x ; \texttt{passage}: p_{m}^{\prime} \right] \right) 
\end{equation}

\noindent where $[ \cdot ;\cdot ]$ denotes the concatenation operation. Each encoded representation $e_m$ is concatenated into a long sequence and fed through the decoder of the generator ($\mathcal{G}_D$), generating the answer sequence $\hat{o}$ token-by-token:
\begin{equation}
\hat{o}=\mathcal{G}_D\left( \left[ e_1;e_2;\dots ;e_k \right] \right) 
\end{equation}

\noindent The generator is optimized by minimizing the cross-entropy loss between the predicted answer $\hat{o}$ and the golden answer $o$ in training.

\section{Experimental Settings}

\subsection{Datasets}

\begin{table}
\centering
\caption{The number of samples in the dataset standardized by the KILT benchmark.}
\resizebox{1.\linewidth}{!}{
\begin{tabular}{l ccc}
    \hline
    \hline

    \textbf{Dataset} & \textbf{Train} & \textbf{Dev} & \textbf{Test} \\ 

    \hline

    \emph{Zero-Shot Slot Filling} & ~ & ~ & ~  \\
    Zero Shot Relation Extraction (ZSRE) \cite{DBLP:conf/conll/LevySCZ17} & 147,909 & 3,724 & 4,966 \\

    \hline

    \emph{Open-Domain Question Answering} & ~ & ~ & ~ \\
    TriviaQA \cite{DBLP:conf/acl/JoshiCWZ17} & 61,844 & 5,359 & 6,586 \\
    Natural Questions (NQ) \cite{kwiatkowski-etal-2019-natural} & 87,372 & 2,837 & 1,444 \\

    \hline

    \emph{Knowledge-Enhanced Dialogue Generation} & ~ & ~ & ~ \\
    Wizard of Wikipedia (WoW) \cite{dinan2018wizard} & 63,734 & 3,054 & 2,944 \\
    
    \hline
    \hline
\end{tabular}
 }
\label{data_statistics}
\end{table}

\begin{table*}[t]
\centering
\caption{Performance on the KILT test set. We report results on the KILT Leaderboard. \textbf{Bold} and \underline{underline} indicate the best and second best approach respectively. ``\emph{K.-}'' is the KILT score for the corresponding metric. ``\emph{R-P}'' and ``\emph{R@5}'' denotes the R-precision and Recall@5, respectively. All models apart from the KILT-WEB-2 employ BART{\footnotesize{LARGE}} as the backbone of the answer generator.}
\renewcommand\arraystretch{1.7}
\setlength{\tabcolsep}{1.5pt}
\resizebox{1.0\linewidth}{!}{
\begin{tabular}{l | cccc | cccc | cccc | cccc}
    \hline
    \hline

    \multirow{2}*{\textbf{Models}} & 
    \multicolumn{4}{c|}{\emph{Zero-Shot Relation Extraction}} & 
    \multicolumn{4}{c|}{\emph{Wizard of Wikipedia}} & 
    \multicolumn{4}{c|}{\emph{Natural Questions}} & 
    \multicolumn{4}{c}{\emph{TriviaQA}} \\

    \cline{2-17}
    
    ~ & \textbf{R-P} & \textbf{R@5} & \textbf{ACC (K.-)} & \textbf{F1 (K.-)} &  
    \textbf{R-P} & \textbf{R@5} & \textbf{R-L (K.-)} & \textbf{F1 (K.-)} &
    \textbf{R-P} & \textbf{R@5} & \textbf{EM (K.-)} & \textbf{F1 (K.-)} &
    \textbf{R-P} & \textbf{R@5} & \textbf{EM (K.-)} & \textbf{F1 (K.-)} \\
    
    \hline

    KGI$_0$ & 94.2 & 95.2 & 69.0 (68.3) & 74.5 (73.5) &
    55.4 & 78.5 & 16.4 (10.4) & 18.6 (11.8) &
    63.7 & 70.2 & 45.2 (36.4) & 53.4 (41.8) &
    60.5 & 63.5 & 61.0 (42.9) & 66.6 (46.1) \\

    KGI$_1$ & \underline{98.5} & 99.2 & 72.6 (72.3) & 77.1 (76.7) & - & - & - & - & - & - & - & - & - & - & - & - \\

    KILT-WEB-2 & 89.6 & 97.9 & 74.0 (67.2) & 78.4 (71.0) &
    41.5 & 68.3 & 13.9 (6.6) & 15.7 (7.6) & 
    59.8 & 71.2 & 51.6 (35.3) & 60.8 (40.7) & 
    58.9 & 71.6 & 72.7 (45.6) & 79.5 (49.6) \\

    HindSight & - & - & - & - &
    56.1 & 74.3 & \underline{17.1 (11.9)} & \underline{19.2 (13.4)} &
    - & - & - & - &
    - & - & - & - \\

    SEAL & 98.0 & \underline{99.3} & \underline{74.6 (73.2)} & \underline{79.7 (78.1)} &
    57.6 & 79.0 & 16.7 (10.5) & 18.3 (11.6) &
    63.2 & 68.2 & \underline{53.7} (38.8) & \underline{62.2} (44.4) & 
    68.4 & \underline{76.4} & 70.9 (50.1) & 77.3 (55.0) \\

    Re$^2$G & - & - & - & - &
    \underline{60.1} & \textbf{80.0} & 16.8 (11.4) & 18.9 (13.0) &
    \textbf{70.8} & \underline{76.6} & 51.7 (\textbf{43.6}) & 61.0 (\underline{49.8}) & 
    \underline{72.7} & 74.2 & \underline{76.3 (57.9)} & \underline{81.4 (61.8)} \\

    \hline
    
    \textbf{\textsc{GripRank}} & \textbf{99.3} & \textbf{99.7} & \textbf{74.7 (74.3)} & \textbf{80.3 (79.9)} & \textbf{63.5} & \underline{79.3} & \textbf{18.1 (12.9)} & \textbf{20.5 (14.7)} & \underline{70.2} & \textbf{77.6} & \textbf{54.0 (43.6)} & \textbf{63.2 (50.3)} & \textbf{73.0} & \textbf{78.8} & \textbf{77.9} (\textbf{58.2)} & \textbf{83.3} (\textbf{62.4)}   \\
    
    \hline
    \hline
\end{tabular}
}
\label{kilt_leaderboard}
\end{table*}

We conduct our experiment on four datasets, across three tasks in the KILT benchmark, including zero-shot slot filling, open-domain question answering, and knowledge-enhanced dialogue generation.

\begin{itemize}

    \item \textbf{Zero-Shot Slot Filling} We use the ZSRE (Zero-Shot Relation Extraction) \cite{DBLP:conf/conll/LevySCZ17} to evaluate our method on the zero-shot slot filling. The ZSRE is originally designed for relation extraction. The KILT casts it as a zero-shot slot-filling task, where a head entity and a relation are regarded as input for the system and the system is expected to output the tail entity for slot-filling.

    \item \textbf{Open-Domain Question Answering} We use the TriviaQA \cite{DBLP:conf/acl/JoshiCWZ17} and NQ (Natural Questions) \cite{kwiatkowski-etal-2019-natural} to evaluate our method on the open-domain question-answering task. Both datasets are a collection of question-answer-evidence triples, and the relevant Wikipedia page for each sample can be explicitly found by a retrieval step.

    \item \textbf{Knowledge-Enhanced Dialogue Generation} We use the Wizard of Wikipedia (WoW) \cite{dinan2018wizard} to evaluate our method on the knowledge-enhanced dialogue generation task. The input for this task is a dialogue history ending with the information seeker's turn.  The output is a response to the mentioned content grounded on a relevant Wikipedia page.
    
\end{itemize}

The statistic of each dataset is shown in Table \ref{data_statistics}.
Unlike the previous work \cite{DBLP:journals/corr/abs-2209-14290,DBLP:journals/corr/abs-2207-03030} that trains the model on the training and passage sets filtered from the original dataset.
We train our model on the dataset standardized by the KILT, facilitating a fair comparison with other work.

\subsection{Evaluation Metrics}
For all tasks in KILT, a system is expected to correctly retrieve the provenance of golden answers from the KILT knowledge sources, and then ground on the retrieved evidence to guide the generation of the answer.
Following previous work \cite{DBLP:conf/naacl/PetroniPFLYCTJK21,maillard-etal-2021-multi}, we use R-Precision (R-Prec) and Recall@5 (R@5) to measure the correctness of the retrieved provenance, where R-Prec and Recall@5 figure out how many retrieved results meet that ground-truth provenance appears in the top-1 and top-5 passages retrieved by the system, respectively.
We use the F1 score to measure the correctness of the generated output at the unigram level for all tasks.
In addition, we use accuracy (ACC), Rouge-L (R-L) \cite{rouge2004package}, and exact match (EM) as indicative metrics to evaluate the specific task accordingly.
Moreover, we use the KILT version of these metrics \cite{DBLP:conf/naacl/PetroniPFLYCTJK21} to indicate the correctness of the output when provenance is ranked at the top in predictions.

\subsection{Baselines}
We compare our approach with a set of competitive baselines:

\begin{itemize}
    
    \item \textbf{KGI} \cite{DBLP:conf/emnlp/GlassRCG21} trains the DPR and the RAG models in a task-specific manner. They constructed hard negatives by fetching the candidate passages from the pre-trained DPR, which boosts the retrieval results by a significant margin and, subsequently, the performance of downstream tasks.

    \item \textbf{KILT-WEB 2} \cite{DBLP:journals/corr/abs-2112-09924} created a new knowledge source \textsc{Sphere} and conducted experiment over it. 
    They employ the conventional retrieve-then-generate approach to achieve the KILT tasks, where BM25 or DPR is employed first to retrieve relevant passages. Then they use FiD \cite{DBLP:conf/eacl/IzacardG21} as a reader component to produce answers.

    \item \textbf{Hindsight} \cite{DBLP:conf/iclr/ParanjapeKPZM22} proposed leveraging an additional retriever that can approach the posterior information to guide the training of the passage retriever, which achieves competitive performance on the WoW dataset.

    \item \textbf{SEAL} \cite{NEURIPS2022_cd88d62a} introduced a novel generative approach for text retrieval. It can generate any strings that appeared in the dataset with the FM-index \cite{ferragina2000opportunistic}, and then map the generated contents to passages.

    \item \textbf{Re$^2$G} \cite{DBLP:conf/naacl/GlassRCNCG22} employed a passage re-ranker with cross-encoder architecture to improve the ranking quality of the candidate passages retrieved by the passage retriever. They achieved state-of-the-art performance on four tasks over the KILT benchmark.
    
\end{itemize}

\subsection{Implementation Details}

\subsubsection{Model Settings}

The passage retriever is in a dual-encoder architecture.
Both encoders use BERT{\footnotesize{BASE}} as the backbone, each with 110M parameters.
We initialize the passage retriever from Dense Passage Retriever (DPR)  \cite{DBLP:conf/emnlp/KarpukhinOMLWEC20} from HuggingFace \cite{DBLP:conf/emnlp/WolfDSCDMCRLFDS20}, and train it with hard negatives fetched from BM25 \cite{DBLP:journals/ftir/RobertsonZ09}, similar to the implementation in \citet{DBLP:conf/emnlp/GlassRCG21,DBLP:conf/naacl/GlassRCNCG22}.
The passage ranker is in a cross-encoder architecture, which is composed of  a single Transformer encoder enabling interactively models of queries and passages.
We use ELECTRA{\footnotesize{BASE}} (110M) pre-trained on MS MARCO dataset \cite{DBLP:conf/nips/NguyenRSGTMD16} as the backbone of the passage ranker.
Both the generative passage estimator (GPE) and the answer generator are trained to do sequence-to-sequence generation with encoder-decoder architecture, which is initialized from BART{\footnotesize{LARGE}} (406M) on HuggingFace.

\subsubsection{Hyper-parameter Settings}

There are four components in our model, i.e., 
passage retriever, passage estimator, passage ranker, and answer generator.
We set the learning rate to 5e-5 for the passage retriever, and 3e-5 for the other three components.
We use a linear learning rate scheduler for all components, with 10\% of the total steps for warmup updates.
The batch size is set to 128 for the passage retriever and the answer generator, and 32 for the passage estimator and the passage ranker.
We train all components with up to 5 epochs.
Moreover, we set the scale of the warm-up passage pool $N_0$, the warm-up steps $T_0$, and the curriculum steps $T$ to 5, 500, and 1000 for all datasets during passage ranking, respectively.
During the inference stage, we set the number of passages utilized for the answer generator to 15 and the beam size to 3.
We use Pytorch \cite{DBLP:conf/nips/PaszkeGMLBCKLGA19} and Huggingface \cite{DBLP:journals/corr/abs-1910-03771} to implement our model, and conduct training and inference on four Tesla V100 machines.


\section{Experimental Results}

We conduct extensive experiments to answer the following research questions:

\begin{itemize}
    \item \textbf{RQ1:} How does the proposed \textsc{GripRank} perform compared with the state-of-the-art retrieval-enhanced text generation models?
    \item \textbf{RQ2:} Is the proposed \textsc{GripRank} really narrowing the gap between retrieval and generation?
    \item \textbf{RQ3:} Is the proposed Curriculum Knowledge Distillation beneficial to help the ranker distinguish the provenance from many plausible candidates?
    \item \textbf{RQ4:} How do different generative language models impact the distillation quality of the generative passage estimator?
    \item \textbf{RQ5:} Can we better understand how different models perform via some case studies?
    
\end{itemize}

\subsection{Performance Comparison}

To answer \textbf{RQ1}, we report the results of each model on ZSRE (zero-shot slot filling task), WoW (knowledge-enhanced dialogue generation task), NQ and TriviaQA (open-domain question answering) in Table \ref{kilt_leaderboard}.

Generally, the proposed \textsc{GripRank} outperforms the previous state-of-the-art methods for both passage ranking and answer generation.
For example, compared to SEAL, one of the strongest baselines on these four datasets, the \textsc{GripRank} achieves around 7.3\% and 4.7\% relative improvements on average according to R-precision and Recall@5 respectively, indicating the superior performance of our method on passage ranking.
With regard to answer generation, the \textsc{GripRank} improves SEAL by around 5.5\% and 13.9\% according to F1 and KILT-F1 on average respectively, indicating that the better passage ranking performance of \textsc{GripRank} further improves the quality of generated answers.
The above observation is still consistent in contrast to the best baselines on each dataset according to the corresponding metric.
For example, the \textsc{GripRank} outperforms the previous state-of-the-art on each dataset according to R-precision and F1 by around 1.5\% and 2.9\% relative improvements on average, respectively.

Interestingly, compared with other baselines, we observe that the \textsc{GripRank} shows potential in correctly predicting the answer no matter if the golden passage ranks at the top.
For example, on the ZSRE dataset, the \textsc{GripRank} outperforms one of the strongest baselines SEAL by around 1.1 and 1.8 points in terms of KILT-ACC and KILT-F1 respectively, even though they perform comparably according to ACC and F1 score.
While on the NQ and TriviaQA dataset, the \textsc{GripRank} outperforms one of the strongest baselines Re$^2$G by around 2.0 and 2.1 points on average in terms of EM and F1 respectively, even though they perform comparably according to the counterpart of the KILT metrics. 
It indicates the superiority of the proposed ranking strategy, which enhances the answer generation by making better use of the golden passage with the help of other top-ranked candidates for generating correct answers.

\subsection{Effect of Passage Ranking}

\begin{table*}[t]
\centering
\caption{Ablation results on the KILT validation set.}
\begin{tabular}{l|ccc|ccc|ccc|ccc}
    \hline
    \hline

    \multirow{2}*{Models} & 
    \multicolumn{3}{c|}{ZSRE} &
    \multicolumn{3}{c|}{WoW} &
    \multicolumn{3}{c|}{TriviaQA} &
    \multicolumn{3}{c}{NQ} \\
    
    \cline{2-4}
    \cline{5-7}
    \cline{8-10}
    \cline{11-13}
    
    ~ & R-prec & ACC & KILT-ACC
    & R-prec & F1 & KILT-F1
    & R-prec & EM & KILT-EM
    & R-prec & EM & KILT-EM \\

    \hline

    \textbf{\textsc{GripRank}} & \underline{99.0} & \textbf{73.8} & \textbf{73.5} & \textbf{58.5} & \textbf{20.2} & \textbf{14.3} & \underline{74.3} & \textbf{76.8} & \textbf{58.2} & \underline{72.0} & \textbf{57.3} & \textbf{46.4}  \\

    ~ w/o Ranker & 92.8 & 66.4 & 65.0 & 51.2 & 18.6 & 12.8 & 62.2 & 61.7 & 45.6 & 66.0 & 50.1 & 39.5 \\

    ~ w/o Distill & 98.2 & 72.0  & 71.5 & 57.0 & 19.2 & 13.4 & 73.8 & 74.3 & 56.8 & 71.6 & 55.7 & 45.1  \\

    ~ w/ Encoder Distill & \textbf{99.4} & 72.5 & 72.3 & \underline{57.6} & \underline{19.4} & \underline{13.6} & \textbf{74.8} & 75.2 & 57.4 & \textbf{72.3} & 56.1 & 45.6 \\

    ~ w/ KL Distill & 98.6 & \underline{73.0} & \underline{72.5} & 57.0 & \underline{19.4} & 13.4 & 73.5 & \underline{75.6} & \underline{58.0} & 71.0 & \underline{56.4} & \underline{45.9} \\

    \hline
    \hline
\end{tabular}
\label{ablation_gpe}
\end{table*}

To answer \textbf{RQ2}, we consider the following variants to conduct ablation:
\textbf{i)} \emph{w/o Ranker}: We remove the passage ranker from the proposed \textsc{GripRank}. Thus, only the top-ranked passages fetched from the passage retriever (i.e., DPR) are used to guide the answer generation.
\textbf{ii)} \emph{w/o Distill}: We only use NLL loss (defined in Eq. \ref{nll_loss}) to supervise the training of the passage ranker without distilling knowledge from teacher models.
\textbf{iii)} \emph{w/ Encoder Distill}: We replace the generative passage estimator (GPE) with an encoder-only teacher model, which shares the same architecture with the passage ranker but additionally considers the golden answer as input. The student passage ranker is under posterior-guided training from the teacher model.
\textbf{iv)} \emph{w/ KL Distill}: We replace the \emph{ListMLE} loss defined in equation \ref{listmle_loss} with \emph{Kullback-Leibler Divergence} loss (KLD loss), supervising the passage ranker to learn the output distribution of the GPE.

Table \ref{ablation_gpe} shows the ablation results, from which we have the following observations:
First, the passage ranker plays a significant role in both passage retrieval and answer generation. 
Results over all of the metrics degrade significantly when the passage ranker is removed from the proposed approach.
Second, all distillation strategies are beneficial to answer generation.
For example, the variant \emph{w/ Encoder Distill}  and the variant \emph{w/ KL Distill} outperform the baseline without distillation (i.e., \emph{w/o Distill}) in terms of the metric to measure generated answers (e.g., ACC, F1, and EM).
Third, the proposed \textsc{GripRank} outperforms the variant \emph{w/ KL Distill}, indicating that the ranker trained by distilling the order knowledge from the GPE outperforms its counterpart distilling whole output distribution.
We speculate that the output distribution of GPE may be too complex for the ranker to learn accurately.

More importantly, distilling the knowledge from GPE to the ranker narrows the gap between retrieval and generation.
As shown in Table \ref{ablation_gpe}, the proposed \textsc{GripRank} outperforms one of the competitive variants \emph{w/ Encoder Distill} according to the metrics to evaluate answer generation (e.g., EM and KILT-EM for open-domain question answering), despite the fact that the latter presents a superior performance in passage retrieval
in terms of \emph{R-prec}.
It indicates that simply improving the ranking capability of the ranker will not be beneficial to narrow the gap between retrieval and generation without explicitly considering how likely the candidate passages can be used to generate the target answer.

\subsection{Effect of Curriculum Knowledge Distillation}

\begin{table*}[ht]
\centering
\caption{Impact of using different generative language models as the generative passage estimator (GPE).}
\resizebox{1.\linewidth}{!}{
\begin{tabular}{l |c|ccc|ccc|ccc|ccc}
    \hline
    \hline

    \multirow{2}*{GPEs} & 
    \multirow{2}*{Params} & 
    \multicolumn{3}{c|}{ZSRE} &
    \multicolumn{3}{c|}{WoW} &
    \multicolumn{3}{c|}{TriviaQA} &
    \multicolumn{3}{c}{NQ} \\
    
    \cline{3-5}
    \cline{6-8}
    \cline{9-11}
    \cline{12-14}
    
    ~ & ~ & R-prec & ACC & KILT-ACC
    & R-prec & F1 & KILT-F1
    & R-prec & EM & KILT-EM
    & R-prec & EM & KILT-EM \\

    \hline

    GPT2-medium & 345M & 97.2 & 72.7 & 72.4 & 57.1 & 19.5 & 13.7 & 73.4 & 75.3 & 57.0 & 69.8 & 55.7 & 45.2 \\

    GPT2-large & 774M & 97.7 & 73.2 & 72.8 & 57.6 & 19.9 & 14.0 & 73.8 & 76.0 & \underline{58.0} & 70.5 & 56.1 & 45.5 \\

    T5-base & 220M & 98.0 & 73.0 & 72.7 & 57.4 & 19.8 & 14.2 & 73.5 & 75.9 & 57.8 & 70.8 & 56.6 & 45.7 \\

    T5-large & 770M & \underline{98.6} & \textbf{74.4} & \textbf{74.2} & \underline{58.0} & \textbf{20.5} & \textbf{14.6} & \textbf{74.3} & \underline{76.4} & 57.9 & \underline{71.5} & \underline{56.9} & \underline{46.3} \\

    BART-base & 139M & 97.8 & 72.9 & 72.6 & 57.1 & 19.6 & 13.8 & 73.2 & 75.5 & 57.5 & 71.1 & 56.5 & 45.9 \\

    BART-large & 406M & \textbf{99.0} & \underline{73.8} & \underline{73.5} & \textbf{58.5} & \underline{20.2} & \underline{14.3} & \textbf{74.3} & \textbf{76.8} & \textbf{58.2} & \textbf{72.0} & \textbf{57.3} & \textbf{46.4}  \\

    \hline
    \hline
\end{tabular}
}
\label{impact_gpes}
\end{table*}

To answer \textbf{RQ3}, we use different strategies to fetch $n$ candidates\footnote{We set $n=5$ for all fetching strategies to facilitate a fair comparison.} from the 100 most relevant passages collected by the passage retriever for training the ranker.
We consider three fetching strategies:
\textbf{i)} \emph{Random:} We randomly select $n$ candidates from the top 100 passages retrieved by the passage retriever.
\textbf{ii)} \emph{TopN:} We select $n$ most relevant passages ranked at the top by the passage retriever.
\textbf{iii)}  \emph{CPS:} We sample $n$ candidates by our proposed \underline{C}urriculum \underline{P}assage \underline{S}ampling strategy.
Furthermore, we investigate how each strategy impact passage ranking in distinguishing the relevant passages from many plausible candidates by progressively expanding the candidate passage pool during validation.
We evaluate the performance of each variant through two indicative metrics for passage ranking, i.e., \emph{R-prec} and \emph{Recall@5}.

\begin{figure}[t]
\centering
\includegraphics[width=9.8cm]{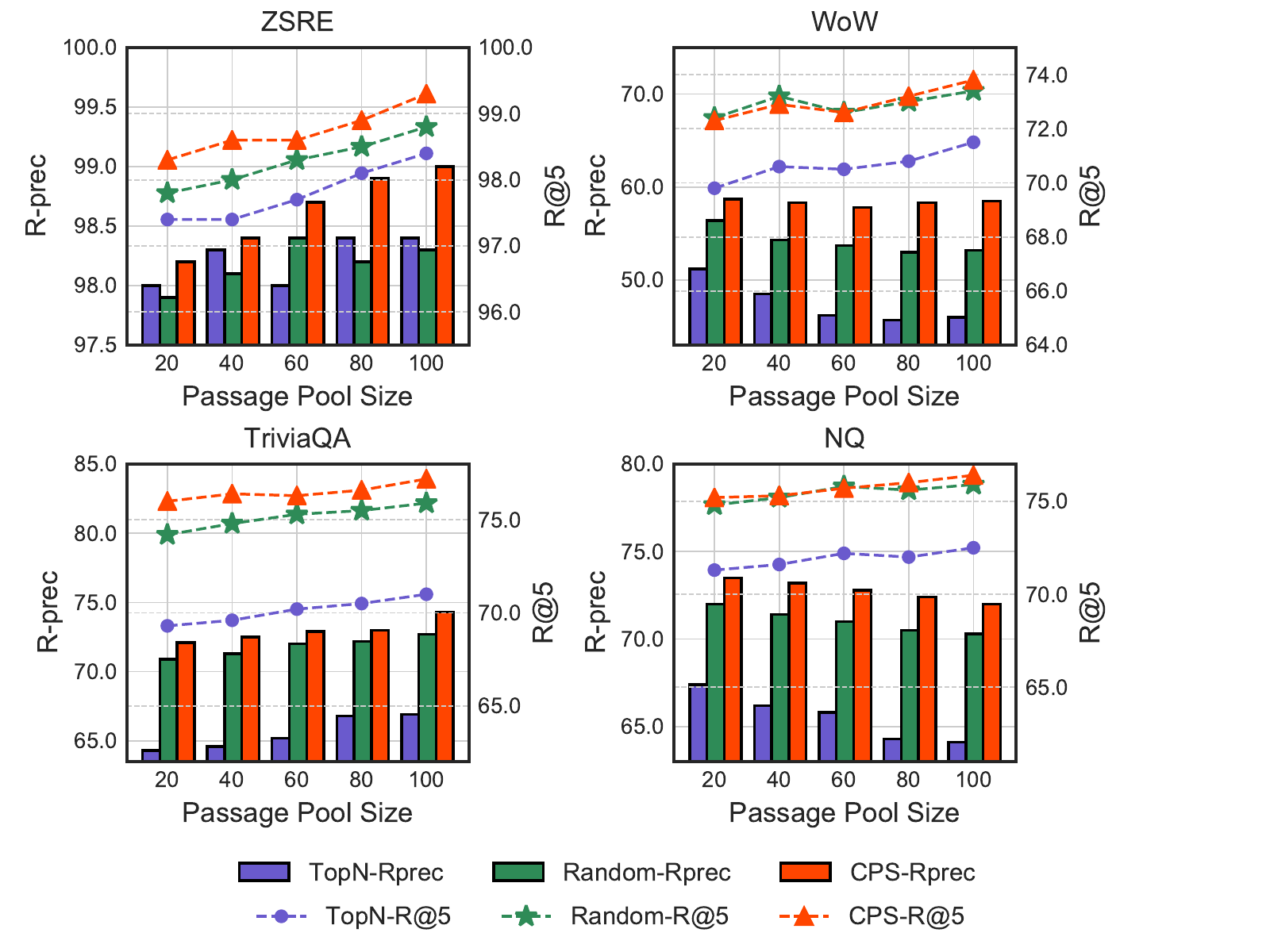}
\caption{The impact of different passage fetching strategies on the passage ranker. We report \emph{R-prec} and \emph{Recall@5} on the validation set of each dataset.} 
\label{effect_cps}
\end{figure}

Figure \ref{effect_cps} shows the impact of different passage fetching strategies on the passage ranker, from which we have the following observations:
i): All variants performed on WoW and NQ datasets lose the retrieval accuracy with the growing size of the candidate passage pool in terms of \emph{R-prec}.
It indicates there are a growing number of plausible candidates with the expansion of the candidate passage pool, challenging the passage ranker to distinguish the provenance from them correctly. 
More importantly, the proposed \emph{CPS} drops the most slightly compared with two other variants, indicating the effectiveness of our proposed approach in distinguishing the answer provenance from many plausible candidates.
ii): Both \emph{CPS} and \emph{Random} fetching strategies outperform \emph{TopN} fetching strategy according to \emph{R-prec} and \emph{Recall@5}, which indicates that simply fetching the top-ranked candidates as hard negatives may be too hard for the passage ranker to distinguish the answer provenance, thus hinder the learning of the passage ranker.
iii): The proposed \emph{CPS} outperforms \emph{Random} fetching strategy across all datasets according to \emph{R-prec}, indicating the effectiveness of the proposed curricula in boosting the ranking capability of the passage ranker.

\begin{table*}[ht]\small
\renewcommand
\arraystretch{1.2}
\centering
\caption{Cases from the ZSRE and the WoW validation set. Blue boldface is the essential associated with the golden answer.}
\resizebox{1.\linewidth}{!}{
\begin{tabular}{r | p{200pt} | p{200pt}}  
    \hline
    \hline

    ~ & \emph{Knowledge-Enhanced Dialogue Generation (WoW)} & \emph{Zero-Shot Slot Filling (ZSRE)} \\
    
    \hline
    \hline
    
    \emph{Input  Query}: & 
    Do you ever enjoy listening to gospel music? & Badrock [SEP] from fictional universe \\

    \hline
    
    \emph{Answer Provenance}: & 
    Gospel music is a \textcolor{blue}{\textbf{genre of Christian music}}... Gospel music usually has dominant vocals (often with strong use of harmony) with Christian lyrics. \textcolor{blue}{\textbf{Gospel music can be traced to the early 17th century, with roots in the black oral tradition}}... 
    & 
    Being one of the most popular members of Youngblood (both in the fictional \textcolor{blue}{\textbf{Image Universe}} and with real-life comic readers), Badrock starred in more spin-off comics than any other member of the team. Many of Badrock's non-Youngblood adventures... 
     \\
    
    \hline
    \hline
    
    \textbf{KGI$_0$}: & 
    I do! I like the music from the 1940s and 1950s. 
    & 
    The Flintlockers
    \\

    \hline

    \textbf{SEAL}: &
    I do! I love the music that is composed and performed for many purposes, including aesthetic pleasure, religious or ceremonial purposes. 
    &
    Youngblood
    \\

    \hline

    \textbf{\textsc{GripRank}}: & 
    I do! I like the music of the \textcolor{blue}{\textbf{early 17th century, with roots in the black oral tradition}}.
    &
    \textcolor{blue}{\textbf{Image Universe}}
    \\

    \hline

    \textbf{Golden Answer}: & 
    I listen to a bit of \textcolor{blue}{\textbf{Christian music}}, but I don't really like the \textcolor{blue}{\textbf{genre of Gospel music}}. I prefer Christian music with a more popular genre.
    &
    \textcolor{blue}{\textbf{Image Universe}} 
    \\
    
    \hline
    \hline
    
\end{tabular}
}
\label{case_study}
\end{table*}

\subsection{Impact of Different Passage Estimators}

To answer \textbf{RQ4}, we employ different generative pre-trained language models, including BART \cite{DBLP:conf/acl/LewisLGGMLSZ20}, T5 \cite{DBLP:journals/jmlr/RaffelSRLNMZLL20} and GPT2 \cite{radford2019language}, to serve as the generative passage estimators.
Table \ref{impact_gpes} reports the results, from which we have two main observations:
i): To the same backbone model, the larger model has a stronger capability in ranking the candidate passages into the correct order. For example, the BART-Large outperforms BART-Base by around 1.2 points on average over all datasets in terms of R-Prec, which subsequently improves the generation performance further.
ii): To the different backbone models, the BART-Large achieves comparable performance with T5-Large with fewer parameters used.
While both the T5-Large and the BART-Large outperform GPT2-Large on both passage retrieval and answer generation.
We suspect it may be attributed to the difference in the pre-training data and tasks used for these generative language models.




\subsection{Case Study}

To answer \textbf{RQ5}, we take input queries from the WoW validation set (knowledge-enhanced dialogue generation task) and the ZSRE validation set (zero-shot slot filling task), and show the generated results from our proposed \textsc{GripRank} and two other competitive baselines, i.e., KGI$_0$ and SEAL.

As shown in Table \ref{case_study}, we have two main observations:
i): In the case of WoW, the responses generated by KGI$_0$ and SEAL deliver no essentials evidenced by the answer provenance.
In contrast, the \textsc{GripRank} generates a response that is faithful to the provenance.
It incorporates evidence snippets relevant to the dialogue context into the generated response, considering both knowledgeability and conversational ability.
ii): In the case of ZSRE, both KGI$_0$ and SEAL produce incorrect answers. 
Specifically, the KGI$_0$ suffers from the extrinsic hallucination problem \cite{10.1145/3571730}, which generates an answer ``\emph{The Flintlocker}'' that is not mentioned in the provenance.
We suspect it is because the provenance is out-of-recall from the candidates retrieved by KGI$_0$.
In particular, both SEAL and the proposed \textsc{GripRank} deliver an answer that appears in provenance, while the \textsc{GripRank} shows a better ability to make better use of the retrieved passages for generating the correct answer, benefiting from its superior passage ranking capability.

\section{Conclusions}
In this paper, we focus on narrowing the gap between retrieval and generation for retrieval-enhanced text generation methods.
We propose \textsc{GripRank}, a novel approach to improve the passage ranking capability by distilling knowledge from a generative passage estimator to the passage ranker. 
We evaluate our approach on diverse knowledge-intensive language tasks, including zero-shot slot filling, open-domain question answering, and knowledge-enhanced dialogue generation.
Experimental results show that the proposed \textsc{GripRank} presents advantages over previous state-of-the-art approaches.
Further analysis demonstrates the effectiveness of our proposed approach in narrowing the gap between passage retrieval and answer generation.

\begin{acks}
We thank all the anonymous reviewers for their insightful comments. This work was supported in part by the National Natural Science Foundation of China (Grant Nos. 62276017, U1636211, 61672081), the 2022 Tencent Big Travel Rhino-Bird Special Research Program, and the Fund of the State Key Laboratory of Software Development Environment (Grant No. SKLSDE-2021ZX-18).
\end{acks}

\bibliographystyle{ACM-Reference-Format}
\balance
\bibliography{anthology}


\end{document}